\documentclass{llncs}
\usepackage[T1]{fontenc}
\usepackage{graphicx}
\usepackage{hyperref}
\usepackage{color}

\usepackage{amsmath}
\usepackage{amssymb}
\usepackage{booktabs}
\usepackage{subcaption}  

\title{Uncertainty-Aware Remaining Lifespan Prediction from Images}
\author{Tristan Kenneweg \and Philip Kenneweg \and Barbara Hammer}
\institute{University of Bielefeld}
\date{}

\begin{document}
\makeatletter
\renewcommand{\section}{\@startsection{section}{1}{\z@}%
  {-0.8ex \@plus -0.5ex \@minus -0.2ex}%
  {0.5ex \@plus 0.1ex}%
  {\normalfont\large\bfseries}}
\renewcommand{\subsection}{\@startsection{subsection}{2}{\z@}%
  {-0.7ex \@plus -0.3ex \@minus -0.1ex}%
  {0.3ex \@plus 0.1ex}%
  {\normalfont\normalsize\bfseries}}
\makeatother
\maketitle

\begin{abstract}
Predicting mortality-related outcomes from images offers the prospect of accessible, noninvasive, and scalable health screening. We present a method that leverages pretrained vision transformer foundation models to estimate remaining lifespan from facial and whole-body images, alongside robust uncertainty quantification. We show that predictive uncertainty varies systematically with the true remaining lifespan, and that this uncertainty can be effectively modeled by learning a Gaussian distribution for each sample. Our approach achieves state-of-the-art mean absolute error (MAE) of 7.41 years on an established dataset, and further achieves 4.91 and 4.99 years MAE on two new, higher-quality datasets curated and published in this work. Importantly, our models provide calibrated uncertainty estimates, as demonstrated by a bucketed expected calibration error of 0.82 years on the Faces Dataset. While not intended for clinical deployment, these results highlight the potential of extracting medically relevant signals from images. We make all code and datasets available to facilitate further research.
\end{abstract}

\section{Introduction}
In this paper, we address the problem of remaining lifespan prediction from images, which we consider interesting for two major reasons:

First, it is unclear how well this task can be solved in the information-theoretic limit. If we had a perfect predictor, what level of precision could we achieve? Intuitively, one might expect only low maximum precision, since not all information relevant for remaining life prediction is likely to be present in a single image. However, this intuition is challenged by the related task of chronological age prediction. The mean absolute error in the MORPH Album 2 dataset, which provides facial images along with the age of each individual, is reported to be less than 2.5 years \cite{agepred1,agepred2}. In contrast, using DNA methylation markers for the prediction of chronological age, the landmark 2013 paper \cite{horvath2013dna} reported a median error of 3.6 years. In this related task, images demonstrate even greater predictive power than complex DNA methylation measurements. Whether similar success is possible for other medical prediction tasks is unclear.

Second, if high predictive precision can be achieved, this would indicate that substantial health information about an individual can be inferred from images. An atypically low remaining lifespan prediction could serve as an indicator for initiating further medical investigations. Subsequently, predictors that detect specific conditions or estimate laboratory values could be developed. As images are \textit{ubiquitous}, consistently extracting medically relevant data from them would greatly benefit preventive medicine.

In this work, we propose a significantly improved approach to remaining lifespan prediction from images by combining powerful pre-trained vision transformers with a regression head that models prediction uncertainty as a Gaussian distribution. We show that such models produce well-calibrated uncertainty estimates, enabling better interpretation and downstream decision-making. We evaluate all models using 5-fold cross-validation and report mean performance across folds. Our contributions are as follows:
\begin{itemize}
    \item We show that strong predictive performance can be achieved in remaining lifespan prediction tasks using pre-trained foundation models such as DINOv2, without extensive architectural tuning.
    \item We model prediction uncertainty using the Gaussian negative log-likelihood and demonstrate calibration with the bucketed Expected Calibration Error (ECE) adapted to regression.
    \item We curate and publish cleaned and improved versions of an existing mortality dataset, leveraging vision-language models to automatically filter samples.
\item We report state-of-the-art performance with a mean MAE of 4.91 years and a mean calibration error of 0.82 years on the Faces Dataset.
\end{itemize}

Although our approach is not intended for clinical deployment, it underscores the hidden potential of simple images as a first-line screening tool and highlights scalable opportunities for uncertainty-aware health modeling using publicly available visual data.  
The code and datasets are publicly available at: \newline\href{https://github.com/TKenneweg/RLPredictionWithUncertainty}{github.com/TKenneweg/RLPredictionWithUncertainty}.

\section{Related Work}

\textbf{Epigenetic clocks and biological ageing}. Estimating individual ageing rates from molecular data has been a major focus since Horvath's seminal pan‑tissue DNA‑methylation clock \cite{horvath2013dna}. Second‑generation clocks such as DNAm PhenoAge \cite{phenoage} and DNAm GrimAge \cite{lu2019dnagrimage} improve correlations with disease burden and all‑cause mortality, and are now widely used in geroscience studies. Crucially, however, these clocks are optimized for hazard ratios or risk scores rather than for point estimates of the time remaining to an individual, which limits their interpretability for person‑centric counseling.

\textbf{Image‑based age and risk prediction}. Computer‑vision research on facial images initially targeted chronological age regression, achieving mean absolute errors below 2.5 years on the MORPH dataset \cite{agepred1,agepred2}. Fekrazad \cite{fekrazad2023estimating} collected a corpus of 24,000 Wikipedia portraits and trained convolutional models to regress \emph{remaining} lifespan, reporting an MAE of 8.3 years. More recently, Bontempi \emph{et al.} introduced \emph{FaceAge}, a deep model that estimates biological age from a single headshot and enhances survival prediction in oncology cohorts \cite{bontempi2025faceage}. While FaceAge improves survival \emph{hazard} modeling, it does not produce calibrated estimates of absolute years of life remaining.

\textbf{Other imaging modalities}. Beyond faces, retinal fundus photographs \cite{nusinovici2022retinal_age_gap}, optical-coherence-tomography volumes \cite{kim2023octage}, and chest radiographs \cite{oakdenrayner2019cxrrisk,gao2023cxrlungrisk} have also been used to forecast longevity‐related outcomes. Most of these pipelines feed image embeddings from a CNN (or Vision Transformer) into a \emph{Cox proportional-hazards loss}—the classical survival-analysis objective that learns a log-linear function producing a \emph{hazard ratio}—or its deep-learning analogue \emph{DeepSurv}, which replaces the linear layer with an MLP while maintaining the same partial-likelihood formulation. Both objectives optimize the \emph{relative risk of death over time} rather than predicting a concrete number of years left. Consequently, their outputs are survival curves or risk scores, underscoring that hazard modeling addresses a related, but distinct, question from the remaining-lifespan regression tackled in the present work.

\textbf{Uncertainty‑aware prediction}. Reliable confidence estimates are indispensable for clinical deployment. Bayesian neural networks \cite{kendall2018what} and deep ensembles \cite{lakshminarayanan2017simple} have been proposed, but calibration for image-based lifespan prediction remains under‑explored. Our Gaussian mean–variance head follows the mean‑variance estimation paradigm \cite{mve_basis,Seitzer2022Pitfalls} and is, to our knowledge, the first to quantify uncertainty in image‑based lifespan regression.

\section{Dataset}
\begin{figure}
    \centering
    \begin{subfigure}[b]{0.48\textwidth}
        \centering
        \includegraphics[width=\textwidth]{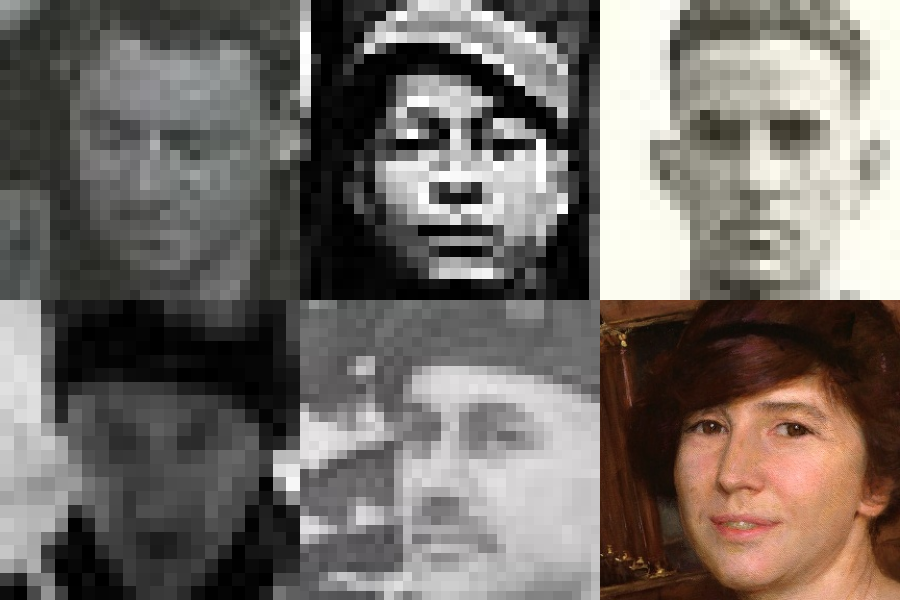}
        \caption{Problematic images from the Legacy Dataset. The image on the bottom right depicts a painting.}
        \label{fig:lowres}
    \end{subfigure}
    \hfill
    \begin{subfigure}[b]{0.48\textwidth}
        \centering
        \includegraphics[width=\textwidth]{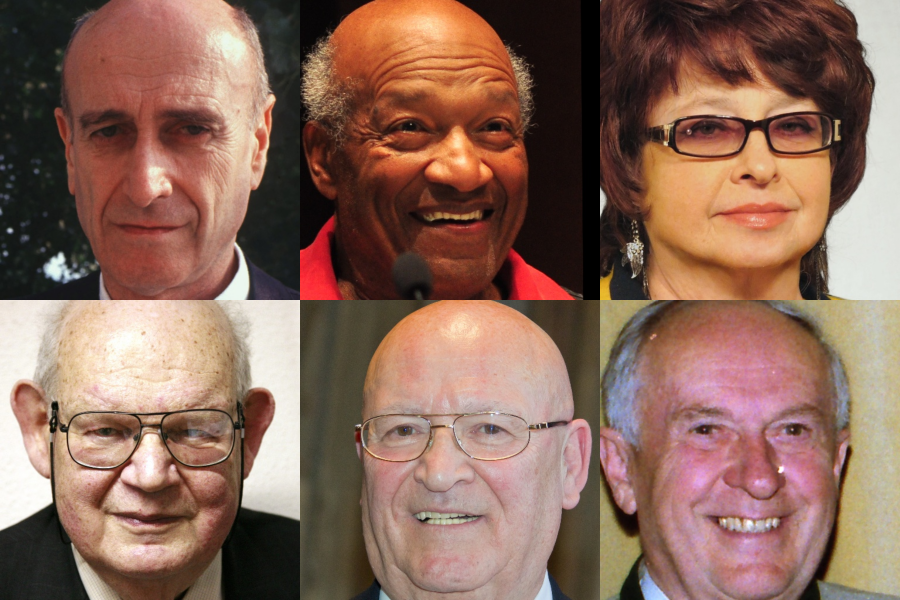}
        \caption{The first six random images of the newly created Faces Dataset.\newline}
        \label{fig:random_imgs}
    \end{subfigure}
    \caption{Dataset Images}
    \label{fig:datasetimgs} 
\end{figure}

We base our dataset on the one published by Fekrazad \cite{fekrazad2023estimating}, henceforth referred to as the Legacy Dataset. The Legacy Dataset contains facial images of people scraped from Wikipedia and Wikidata who died between 1990 and 2022. When no unnatural cause was given, death was presumed to be from natural causes. For details on the scraping process, we refer to the original paper. While overall well-constructed, we identified several quality issues in the Legacy Dataset:
\begin{itemize}
    \item Many images are of extremely low resolution and do not contain color information. We expect fine details and color to be especially important for the task at hand. Examples of inadequate data points can be seen in Figure \ref{fig:lowres}.
    \item Some of the images are not photographs of real people, but rather drawings, paintings, or other non-photographic representations.
    \item Death dates were recorded as integer years, introducing substantial noise in the remaining lifespan targets.
\end{itemize}

To address these issues, we performed an extensive data cleaning process. We retained data points only if they met the following criteria:
\begin{itemize}
    \item \textbf{Resolution and Color}: The facial image must have a minimum resolution of 200$\times$200 pixels and contain color information.
    \item \textbf{Content}: The image must be a photograph of a real person, not a drawing, painting, or a photo of a sculpture.
    \item \textbf{Availability}: The full version of the image, including more than just the face, must be downloadable from Wikidata.
\end{itemize}

We used the multimodal GPT-4o-mini API to automatically determine whether an image contains color and is a photograph of a real person. Additionally, we queried the Wikidata API to retrieve more precise death dates, recorded as floating-point numbers. Finally, we downloaded the corresponding full images from Wikidata (the Legacy Dataset contains only the cropped faces). After filtering, we retained 5,672 data points, with an average remaining lifespan of 11.57 years and a standard deviation of 11.9 years. During manual inspection of the first 200 images, we found no incorrectly included images, indicating that GPT-4o-mini was able to reliably filter non-photographic samples.

We created two datasets: a \textit{Faces Dataset}, which contains the cropped facial images, and a \textit{Whole Images Dataset}, which contains the full Wikidata images, usually depicting part or most of the body. We also retain the \textit{Legacy Dataset}, which contains 24,167 data points, with a mean remaining lifespan of 29.99 years and a standard deviation of 22.04 years. For comparison, we also report results on the Legacy Dataset. 

\begin{figure}
\centering
\includegraphics[width=1\textwidth]{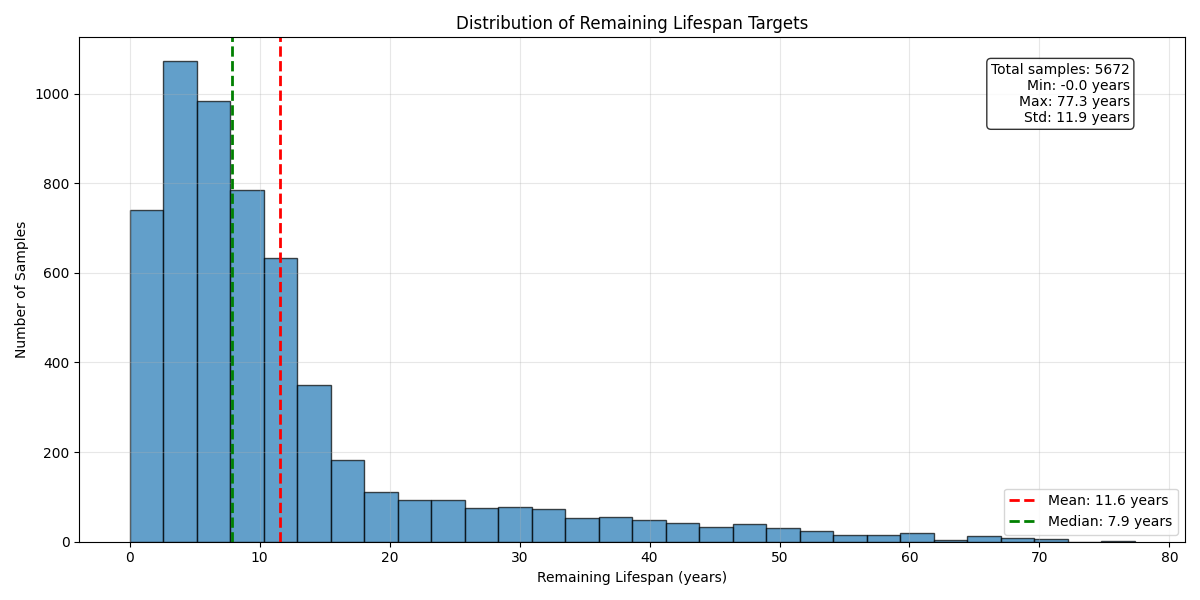}
\caption{Histogram of target values in the newly created datasets.}
\label{fig:target_hist}
\end{figure}

Figure \ref{fig:target_hist} shows the target remaining lifespan distribution in the newly created dataset. We observe a high prevalence of relatively small values, which is expected when scraping images from Wikipedia, since many images have been updated relatively recently prior to or following a person’s death.

\section{Methodology}

Remaining lifespan prediction from images is a complex task involving high-dimensional input data, requiring deep neural networks to achieve good performance. However, we are working with a relatively small dataset that cannot be used to train a deep network from scratch. Thus, we turn to large pre-trained vision transformers, which we use to produce semantically powerful embeddings that can be further processed by a regression head. We compared the performance of three possible backbones, without fine-tuning: DINOv2 \cite{oquab2024dinov2learningrobustvisual} (with learned register tokens), I-JEPA \cite{ijepa}, and CoAtNet \cite{dai2021coatnetmarryingconvolutionattention}. The largest version of DINOv2, with more than 1 billion parameters, yielded clearly superior performance and was therefore used in all subsequent experiments.   

The data in remaining lifespan prediction exhibits heteroskedasticity, meaning that the larger the true or predicted lifespan, the higher the expected error. This is not a failure of the underlying model, but rather an inherent uncertainty in the problem. If a person has a larger remaining lifespan, more factors can potentially influence longevity over time—for example, changes in body composition. The mean absolute error as a metric can thus be misleading, as the uncertainty for a single data point may vary greatly depending on the actual prediction. To mitigate this issue, we constructed a regression head with two output layers: one predicting the mean value $\mu$ (the remaining lifespan prediction), and one predicting the log variance, $\log{\sigma^2}$. We train the regression head using the negative log-likelihood (GNLL) loss for a Gaussian distribution \cite{mve_basis}:
\begin{equation}
\mathcal{L} = \frac{1}{2N} \sum_i\left( \log(\sigma_i^2) + \frac{(y_i - \mu_i)^2}{\sigma_i^2} \right)
\label{eq:gauss}
\end{equation}
where $y_i$ denotes the actual remaining lifespan of the $i$-th data point. This loss encourages the network to predict higher uncertainty when the squared error is large, and vice versa. 
Note that we do not directly predict the expected absolute error, but rather the standard deviation of a Gaussian distribution. The expected absolute error is thus given by
\begin{equation}
    \hat{e_i} = \mathbb{E}\!\bigl[\,|X-\mu|\,\bigr] = \sqrt{\frac{2}{\pi}} \sigma_i \approx 0.798 \sigma_i.
\end{equation}

\subsection{Fine Tuning}
\label{sec:finetune}
\begin{figure}
\centering
\includegraphics[width=1\textwidth]{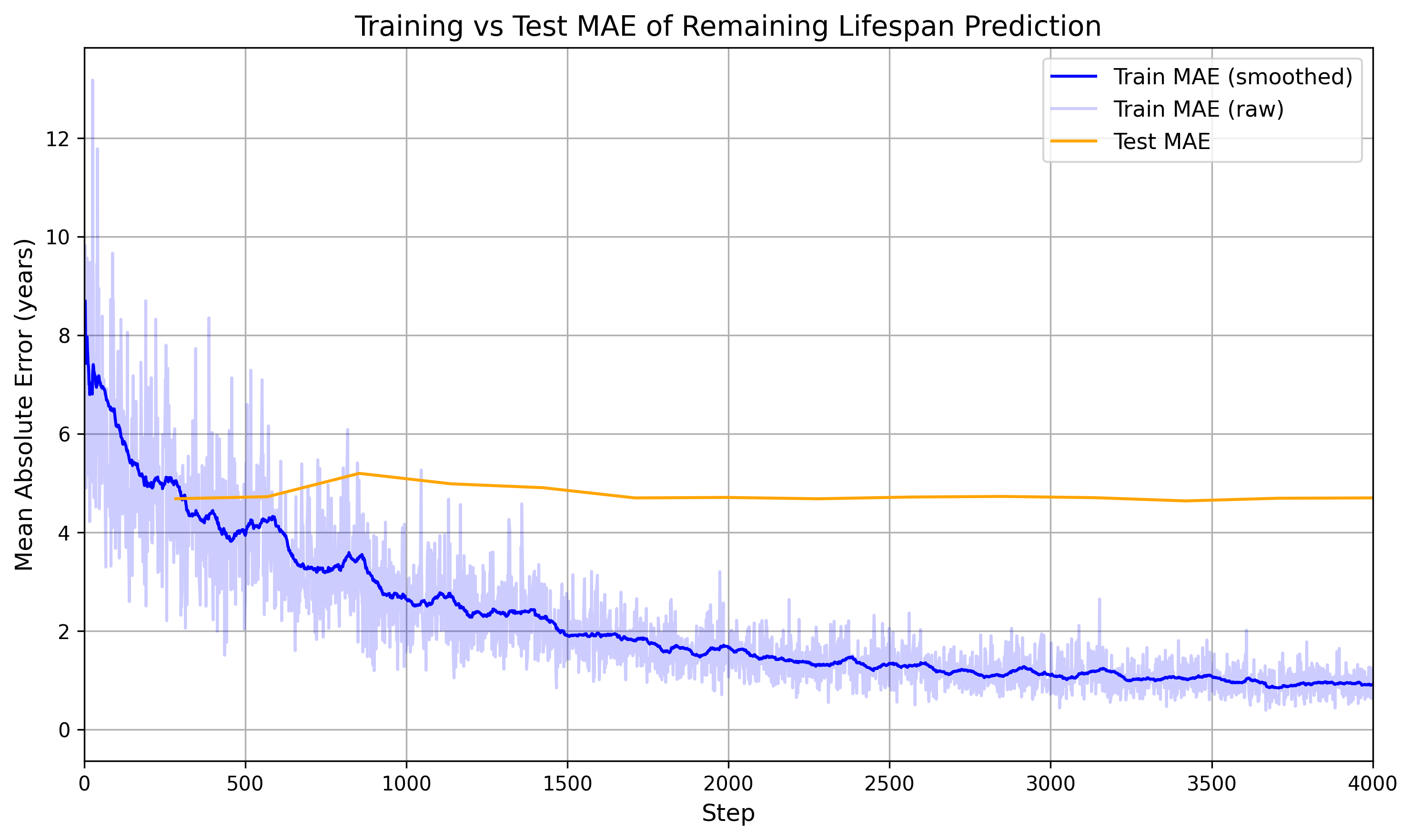}
\caption{Loss curves from an L1-loss training run during which gradients were passed through all layers of the transformer backbone. Note that while the network is overfitting on the training data, this does not result in degraded performance on the test set.}
\label{fig:traintest}
\end{figure}

Overfitting is a major challenge when propagating gradients through the large vision transformer backbone, which has over 1 billion parameters. Multiple techniques exist to mitigate this problem, most notably gradual unfreezing of layers combined with an adaptable learning rate schedule. However, we found that none of these techniques are necessary as long as we choose a suitably low learning rate and train on the L1 loss. The transformer would overfit the training set, but this did not lead to reduced generalization capability, unlike the behavior typically observed in smaller neural networks \cite{jiang2024unveil}. Although the training loss was lower than the test loss, the test loss continued to decrease with each epoch, as seen in Figure \ref{fig:traintest}. 

However, this behavior does not transfer to the GNLL loss. The network produces certainty estimates that are overly optimistic, as the predictions on the training set are very accurate.

To mitigate this issue, we employ a two-phase strategy. First, we train the backbone and regression head on the L1 loss, without back-propagating any gradients from the variance estimation output layer. Second, we freeze the backbone and train only the head on the GNLL loss. However, this approach requires a further split of the training dataset, since the backbone will have overfitted the original training data.

This further split reduces the number of data points available for training the uncertainty estimation, which is especially problematic for remaining lifespan buckets with few samples (see Figure \ref{fig:target_hist}). Since backbone fine-tuning only improved MAE by 0.1–0.2 years over the non-fine-tuned version, we chose to forego the fine-tuning step, preferring better uncertainty estimation at the cost of minor performance losses.

\section{Experiments and Evaluation}

\subsection{Implementation Details}
All training and evaluation runs were performed with a manually set PyTorch random seed (set to 1) to prevent random variations in results. We evaluate using 5-fold cross-validation (KFold with shuffling, random state 1), reporting the mean test performance and standard error across folds.

Our regression head is a small MLP that receives the output of the vision transformer backbone and has two output heads to predict $\mu$ and $\sigma$. These heads share two layers of common parameters, as we observed worsened calibration when using separate parameters. The regression head consumes the prepended "classification" token of the DINOv2 architecture, which has been specifically trained to provide information for downstream image tasks.

We applied a standard normalization procedure to achieve approximately a mean and standard deviation of one for our target remaining lifespan data, and used an image normalization procedure similar to that used during backbone pre-training. For facial images, we resize to 224$\times$224 pixels according to DINOv2 pre-training. For whole-body images, we increase this to 1022$\times$1022, as we expect relevant details to be lost otherwise.

\subsection{Metrics}
We report the negative log-likelihood Gaussian loss (GNLL, see Equation \ref{eq:gauss}), the mean absolute error (MAE), and the bucketed version of the expected calibration error for regression tasks (bucketed ECE). The bucketed ECE is given by

\begin{equation}
    ECE = \frac{1}{N}\sum_{b=1}^{10}{n_b}|e_b - \hat{e_b}|.
\end{equation}

where $n_b$ denotes the number of samples in a given bucket, $e_b$ the true MAE in a given bucket, and $\hat{e_b}$ the predicted mean absolute error. We divide the data into 10 buckets.

We choose the bucketed ECE \cite{stirn2023faithful,zelikman2020crude} as it is one of the most expressive single-number calibration metrics for our task. At one extreme, using a single bucket, the bucketed ECE converges to $ECE_1 = \frac{\sum_{i=1}^N |e_i|}{N} - \frac{\sum_{i=1}^N|\hat{e}_i|}{N}$. However, this metric can obscure systematic calibration errors. For example, if the network is underconfident in the low remaining lifespan regime while being overconfident in the high remaining lifespan regime, the single-bucket error would not reflect this and would remain close to zero. At the other extreme, one could use the point-wise calibration error $ECE_p = \frac{1}{N}\sum_{i=1}^N |e_i - \hat{e}_i|$, but this condition is too strict; if the network could perfectly predict its point-wise error, it would not make any errors at all. The bucketed version of the ECE gives a measure of the network’s ability to predict its errors on \textit{average}, while not being as susceptible to obfuscation by systematic errors.

\subsection{Chronological Baseline}
\label{sec:ca}
We were interested in whether our method could predict remaining lifespan with higher precision than is possible using chronological age (CA) alone. Since CA is a strong predictor of mortality, it is fairly easy to obtain a signal correlated with remaining lifespan, as long as the input allows estimation of CA—which images certainly do.

We compared the predictive performance of CA alone by acquiring population-level death rates from the Human Mortality Database \cite{HMD} and calculated the expected remaining lifespan based on an individual’s age at the time the image was taken and the population-level survival rates. In practice, the mortality data cuts off at an age of 110 years; we set the probability of death to one after this age.

Death rates are available by country, but there is no good match for the uneven country distribution of people with Wikipedia entries. We used the average death rates of the USA between 2000 and 2010 as a substitute. Although more population-specific data would increase the precision of the CA prediction, we still consider this method a meaningful baseline.

\subsection{Results}
\begin{table}[ht]
\centering
\caption{Mean test performance (5-fold cross-validation) and standard error on different datasets and baselines.}
\begin{tabular}{llll}
\toprule
Method / Dataset & MAE $\downarrow$ & ECE $\downarrow$ & NLL  $\downarrow$ \\
\midrule
\textbf{Ours (Faces Only)}      & 4.91 $\pm$ 0.2 & 0.82 $\pm$ 0.23 & -0.216 $\pm$ 0.05 \\
\textbf{Ours (Whole Images)}    & 4.99 $\pm$ 0.22 & 0.96 $\pm$ 0.26 & -0.166 $\pm$ 0.07 \\
\textbf{Ours (Legacy Dataset)}  & 7.41 $\pm$ 0.07 & 0.93 $\pm$ 0.08 & -0.391 $\pm$ 0.02 \\
CA Baseline (Faces Only)    & 7.80 & --   & --     \\
CA Baseline (Legacy Dataset) & 8.62 & --   & --     \\
Fekrazad et al. \cite{fekrazad2023estimating} (Legacy Dataset) & 8.30 & --   & --     \\
\bottomrule
\end{tabular}
\label{tab:results}
\end{table}

\begin{figure}
    \centering
    \begin{subfigure}{1\textwidth}
        \centering
        \includegraphics[width=\textwidth]{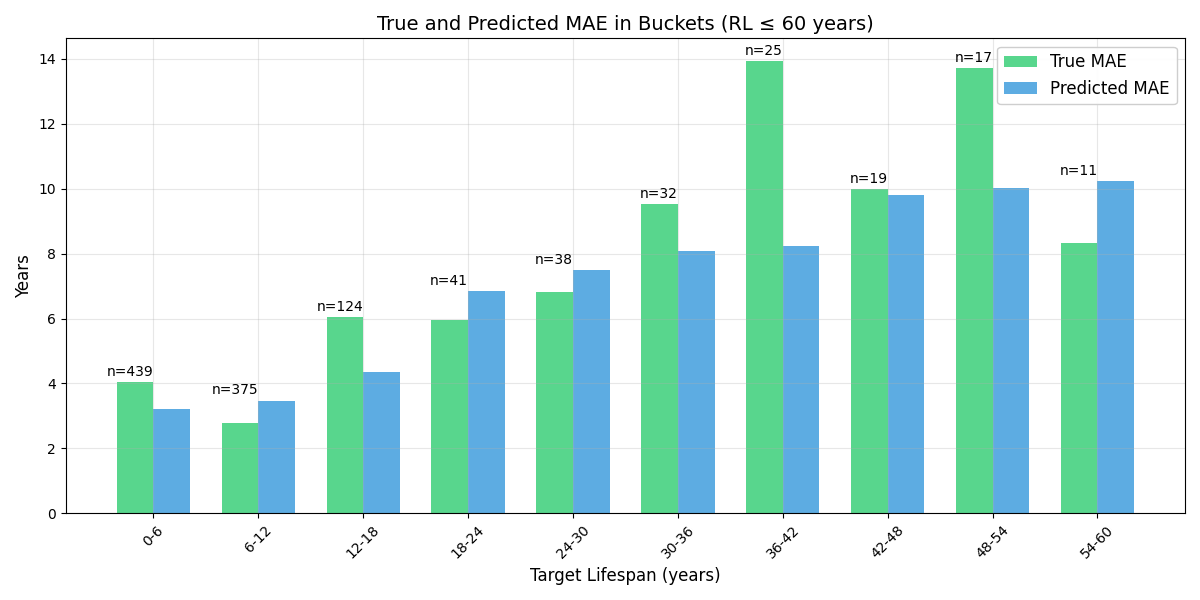}
        \caption{True and predicted errors displayed by bucket up to a remaining lifespan of 60 years.}
        \label{fig:errorbuckets}
    \end{subfigure}
    
    \vspace{1em}

    \begin{subfigure}{1\textwidth}
        \centering
        \includegraphics[width=\textwidth]{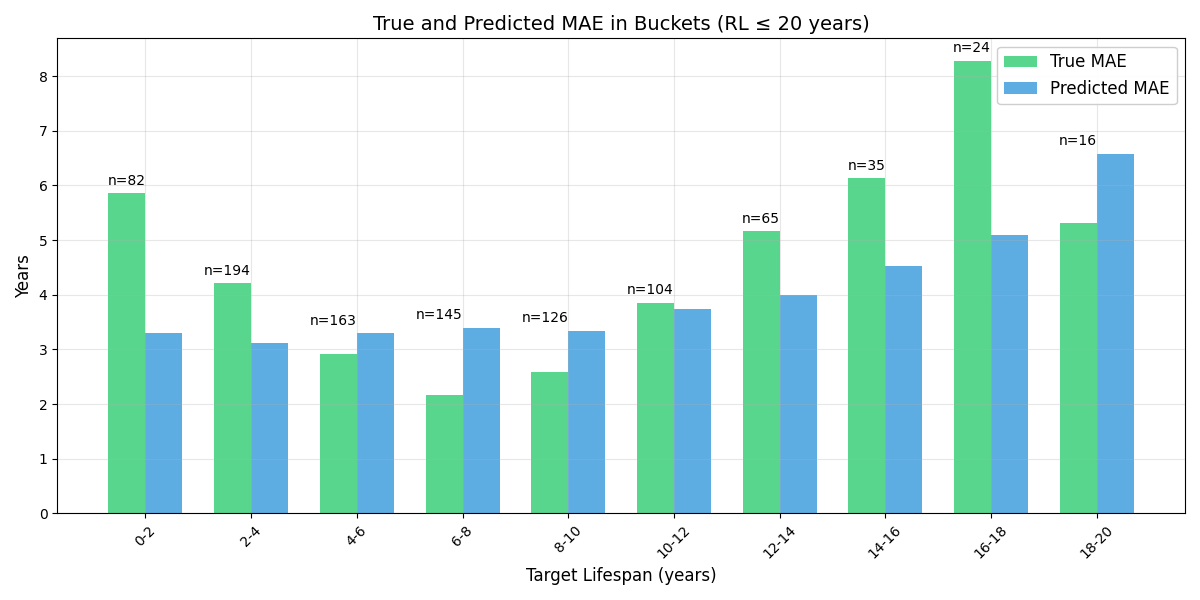}
        \caption{True and predicted errors displayed by bucket up to a remaining lifespan of 20 years.}
        \label{fig:errorbuckets2}
    \end{subfigure}
    
    \caption{Comparison of binned true and predicted errors. (a) shows the error distribution for the whole remaining lifespan range, with a few extreme outliers cut off. (b) Zooms in on the remaining lifespan range of zero to 20 years where most data points lie. The numbers above the bars indicate the number of elements in a given bucket in the test set. The graphs display results on the Faces Dataset.}
    \label{fig:errorbuckets_combined} 
\end{figure}

An overview of the experimental results, including our method, the chronological age (CA) baseline, and the previous state-of-the-art by Fekrazad et al. \cite{fekrazad2023estimating}, is presented in Table~\ref{tab:results}. Our method achieves the best performance across all datasets, substantially reducing mean absolute error compared to both baselines.

Figure~\ref{fig:errorbuckets_combined} provides a detailed analysis of model performance across different remaining lifespan ranges on the Faces Only dataset for the first fold. The comparison between binned true and predicted errors demonstrates that our uncertainty estimates are well-calibrated across most buckets.

\section{Discussion}

\subsection*{Performance}
Our method achieves state-of-the-art performance on remaining lifespan prediction, with a mean MAE of 7.41 years on the Legacy Dataset and substantially better mean results (4.91 and 4.99 years) on our newly curated Faces and Whole Images datasets, respectively (all results via 5-fold cross-validation). However, this improvement is only partly attributable to higher data quality and improved modeling. The new datasets have lower average remaining lifespans, making the prediction task inherently easier—a side effect of filtering for high-quality images, which are more common for recently deceased individuals. Further research should prioritize curating more diverse datasets, for example from newspaper archives, to avoid this bias.

Performance varies across the remaining lifespan spectrum. Our model is most accurate for individuals with a remaining lifespan between 6 and 12 years (MAE = 2.77), while prediction is less reliable for those who die within six years of the photo (MAE = 4.04). The poorest performance in this range occurs for people dying within two years of the photograph, as shown in Figure~\ref{fig:errorbuckets2}. This may reflect either a lack of discriminative visual features or model limitations, and could also be influenced by the presence of unnatural deaths in the dataset (e.g., accidents), which may be overrepresented among people with an an image in their Wikipedia article that was taken close to the end of their life.

We significantly outperform the chronological age baseline, showing that our network uses visual cues that go beyond those which are relevant for chronological age prediction.

\subsection*{Calibration}
The prediction error varies greatly across different remaining lifespan ranges, as expected due to the inherent aleatoric uncertainty of the task. Figure~\ref{fig:errorbuckets_combined} shows that MAE, as a single-number metric, can be misleading when judging the reliability of the network for individual cases. In contrast, the uncertainty estimate produced by the network is well calibrated on average, with a mean bucketed expected calibration error of 0.82 years on the Faces Dataset (95\% CI [0.59, 1.05]).

\subsection*{Facial vs Whole Images}
We consistently observe slightly worse results on the Whole Images Dataset (MAE 4.99 vs. 4.91 on Faces). While most relevant information is expected to be contained in the face, we hypothesized that whole images might provide additional cues (e.g., weight, muscle mass) known to influence longevity, which can only be partially inferred from facial features. We thus conclude that the slight performance gap is inherent to the current approach. Since the DINOv2 backbone compresses the entire image into a single vector of size $d_{emb} = 1536$, we expect it to contain less information about the face when processing whole images. If most relevant information is indeed contained in the face, this would explain the performance gap. However, even when we fine-tuned the backbone using an L1 loss (as described in Section~\ref{sec:finetune}), the performance gap persisted. We specifically provided higher resolution images to preserve facial information and allow the vision transformer backbone to focus on the most relevant parts of the image. We conclude that part of the performance loss is likely due to the pretraining of the DINOv2 model, which was trained on images of 224$\times$224 resolution~\cite{oquab2024dinov2learningrobustvisual}, and does not generalize to larger image resolutions without some loss of performance.

\subsection*{Limitations}
Despite the encouraging results, we emphasize that our work is a proof of concept for the scientific community and not a clinically relevant tool. All data were scraped from Wikipedia and Wikidata, and presumably still contain substantial noise. The remaining lifespan data points are strongly skewed towards shorter values, and it is unclear how well the Wikipedia dataset reflects the general population. We did not test whether multiple images of the same person under different lighting conditions result in consistent remaining lifespan predictions. Furthermore, the model does not account for changes in life expectancy over the decades. Given the same image, we should on average expect a longer remaining lifespan if the photo was taken in 2010 versus 1970. However, including the image date as a feature is problematic, as the network would quickly learn that the dataset does not contain people who died after 2022 and could adjust its predictions accordingly, introducing artifacts. Consequently, a recently taken picture (e.g., from 2025) would be out of distribution.

\section{Conclusion}

We demonstrate that remaining lifespan prediction from facial and whole-body images is feasible and effective using pretrained vision foundation models. We further show that predictive uncertainty strongly depends on the remaining lifespan magnitude, and that this uncertainty can be reliably estimated, achieving a mean bucketed expected calibration error of 0.82 years on the Faces Dataset. Our approach outperforms previous methods on the Legacy Dataset, and we curate and publish two new higher-quality datasets, achieving mean MAEs of 4.91 (Faces) and 4.99 (Whole Images) years, respectively, under 5-fold cross-validation. We hope that our research encourages further investigation into the extraction of medical information from images.

\section*{Acknowledgments}
The authors were supported by SAIL. 
SAIL is funded by the Ministry of Culture and Science of the State of North Rhine-Westphalia under the grant no NW21-059A.

\bibliographystyle{unsrt}
\bibliography{references}

\end{document}